# Nature inspired artificial intelligence based adaptive traffic flow distribution in computer network

Manoj Kumar Singh

**Abstract**— Because of the stochastic nature of traffic requirement matrix, it's very difficult to get the optimal traffic distribution to minimize the delay even with adaptive routing protocol in a fixed connection network where capacity already defined for each link. Hence there is a requirement to define such a method, which could generate the optimal solution very quickly and efficiently. This paper presenting a new concept to provide the adaptive optimal traffic distribution for dynamic condition of traffic matrix using nature based intelligence methods. With the defined load and fixed capacity of links, average delay for packet has minimized with various variations of evolutionary programming and particle swarm optimization. Comparative study has given over their performance in terms of converging speed. Universal approximation capability, the key feature of feed forward neural network has applied to predict the flow distribution on each link to minimize the average delay for a total load available at present on the network. For any variation in the total load, the new flow distribution can be generated by neural network immediately, which could generate minimum delay in the network. With the inclusion of this information, performance of routing protocol will be improved very much.

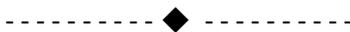

## 1 INTRODUCTION

THE designing of computer network is always a challenging and fascinating task. Because, several domains have been integrated with network, present difficulties are manifold compare to earlier days of network evolution. But characteristics of fundamental problem, determining the most economic way to interconnect nodes while satisfying some reliability and quality of service constraints [1],[2],[3],[4],[5] remain same even today. Two different types of network can be defined [6]:
(1) The centralized network: where server distributes transmission and resource access permission to all other nodes to the network.
(2) The distributed network: where collectively nodes determine the order in which they can send information, by taking into account the multiplicity of available routes.
In the deign of packet switched networks various aspects to be consider to achieve the objectives :(i) the topological configuration that refers to the set of links connecting nodes together, (ii) the traffic that corresponds to the number of packets exchanged per second between each node pair of the network to be designed, (iii) the capacity assignments that consists of determine the maximum number of bits per second (bps) that can be transmitted by each link of a given topological configuration, (iv) the routing scheme that allow selection of the best among the multiple routes connecting each node pair, (v) the flow control procedures, which assure that the quantity of information, sent by the emitter does not overwhelm the receiver.

Quality of network can be defined by several parameters like average delay and reliability of the network [4]. As an index of quality of service, average packet delay in a network can be defined as the mean time taken by a packet to travel from a source to a destination node. The reliability is measured in terms of k-connectivity. Topological design of distributed packet switched networks can be viewed as a search of topologies that minimize communication cost by taking into account delay and reliability constraints. There exist a number of papers, which deal with such approaches [3]. Pierre [8] has proposed the system called SIDRO using artificial intelligence techniques for the topological design of packet switch networks. Mitra et.al [3] have proposed a hybrid method integrating both an algorithmic approach and a heuristic knowledge-based system. Many papers have utilization of simulated annealing. Average transit delay of a packet through the network is one of the most important basic performance criteria. For a given

———————————————
● Manoj Kumar Singh is with the Manuro Tech. Research, Bangalore, India.

network topology of switching nodes and partially connected by communication links; fixed link capacity and fixed traffic requirement; the packet delay is a function of routes employed in forwarding packets in the network. A congested link would be queued and experience excessive waiting time in queue before it can be served (transmitted). Finding the optimal route and hence optimal flow assignment so as to minimize the packet delay has thus become an important issue in the design of packet-switched communication networks [12].

Multipath routing or spatial traffic dispersion [11] is a load balancing technique in which the total load from a source to a destination is partially distributed over several paths. It is useful for relieving congestion and delivering quality of service guarantees in communication networks. Multipath routing has been found to be an effective method to alleviate the adverse effects of traffic burst. In addition, multipath routing protocols helps to spread out congestion and thus minimize network delays. The key to multipath routing is how to allocate a proper portion of traffic to each participating path so as to satisfy the desired objectives, such as the minimization of the average end-to-end path delay. Most of the existing work considered the traffic arrival rate at every link in the network, and found an optimal routing which directs traffic exclusively on least-cost paths with respect to some link costs that depend on the flows carried by the links. It is computationally expensive to find an optimal flow assignment for such source-destination pair. The solution is generally not scalable in terms of the size of the network considered. Adaptive routing schemes have been proposed to spread packets dynamically over multiple paths according to the network load. These procedures require parameters to determine the load distribution. Yet the calculations of such parameters are either computationally intensive or done in an ad-hoc manner. Thus there is need for new multipath routing schemes that allow a rapid computation of the optimal load distribution parameters.

## 2 NETWORK DESCRIPTION: A MODEL APPROACH

A typical distributed computer network can be viewed as a two level hierarchical structure. First level consists of the communication sub network also called backbone. It is comprised of linked switching nodes and has as its main function the end-to-end transportation of information. The second level consists of terminals, workstations, multiplexers, printer and so on. In this paper design related to first level only focus. Each network link is characterized by a set of attributes, which principally are the flow and capacity.

For a given link i, the flow $f_i$ is defined as the effective quantity of information transported by this link, while its capacity $C_i$ is a measure of the maximal quantity of information that it can transmit. Flow and capacity are both expressed in bits/s (bps). Capacity options are only available on the market in discrete or modular options. The traffic $Y_{ij}$ between a node pair (i,j) represents the average number of packets/s sent from source i to destination j. The flow of each link that composes the topological configuration depends on the traffic matrix. Indeed this matrix is varies according to the time, the day and application used.

## 3 PROBLEM DEFINITION

Traffic requirements between nodes arise at random times and the size of the requirement is also a random variable. Consequently, queues of packets build up at the channels and the system behaves as a stochastic network of queues. For routing purposes, packets are distinguished only on the basis of their destination, thus messages having a common destination can be considered as forming a class of customers. The packet switch network therefore can be modeled as a network of queues with 'n' classes of customers where 'n' is the number of different destination.

Average (busy-hour) traffic requirement between nodes can be represented by a requirement matrix R={$\mathbf{r}_{jk}$}, where $\mathbf{r}_{jk}$ is the average transmission rate from Source j to destination k. In some cases we define the requirement matrix as R= ρ k, Where k is a known basic traffic pattern and ρ is a variable scaling factor usually referred to as the traffic level. In general, R (or k) cannot be estimated accurately a priori, because of its dependence upon network parameters (e.g. allocation of resources to computers, demand for resources, etc.), which are difficult to forecast and are subject to changes with time and with network growth. The routing policy and the traffic requirements uniquely determine the vector f ($f_1, f_2, f_3, \ldots f_b$) where $f_i$ is the average data flow on link i.

Because of the continuous dynamic condition of requirement matrix, even for adaptive routing protocol, it's very difficult to determine such a flow vector, which could quickly minimize the average delay of packet in the network with present variation.

## 4 AVERAGE DELAY MODEL IN NETWORK
### 4.1 Delay expression
The average packet delay (T) in a network can be defined as mean time taken by a packet to travel from a source node to a destination node. With the following assumptions (i) external Poisson arrivals (ii) exponential packet length distribution (iii) infinite nodal storage (iv) error free channels (v) no node delay (vi) independence between inter-arrival time and transmission time on each channel, expression for T can be defined as

$$T = (1/\gamma) \sum_{i=1}^{N} f_i / (c_i - f_i) \quad \text{---------------(1)}$$

$$\text{Where } \gamma = \sum_{i=1}^{N} f_i$$

$f_i$ denotes the flow in the link i, $C_i$ is the maximum capacity of link i, $\gamma$ is the total traffic, N is the number of links in the network. More accurate expression for delay can be derived by extending the equation (1), but for most design purpose above equation is accurate.

### 4.2 Optimization model of delay

Optimization model has formulated as Minimization of delay.

To achieve the optimal minimal value of delay, formulation can be defined as

Given:   Topology
         Channel capacities {Ci}
         Requirement matrix R

Minimize:   T

Over the design variable f =(f1, f2, f3…fn)

Subjected to
   a) f is a multicommodity flow satisfying the requirement matrix R

   b) f ≤ C

## 5 OPTIMIZATION USING EC & PSO

Evolutionary computation has experienced a tremendous growth in the last decade in both theoretical analysis and industrial applications. Its scope has evolved beyond its original meaning of biological evolution. Towards a wide variety of nature inspired computational algorithms and techniques, including evolutionary, natural, ecological, social and economical computation etc. in a unified framework, EC is the study of computational system which use ideas and get inspirations from nature evolution and adaption. It is a fast growing interdisciplinary research field in which a variety of techniques and methods are studied for dealing with large complex and dynamic problems. The primary aims of EC are to understand the mechanism of such computational systems and to design highly robust, flexible and efficient algorithms for solving real world problems that are generally very difficult for conventional computing methods. EC was originally divided into four groups: evolutionary strategy (ES), evolutionary programming (EP), genetic algorithm (GA) and genetic programming (GP). Nowadays all the approaches used in EC employ a population based search engine with perturbation (e.g. crossover and mutation) and acceptation (selection and reproduction) to the better solutions compared to conventional optimization methods. The major advantages of EC approaches include: conceptual and computational simplicity, broad applicability, excellent real world problem solvers, potential to use domain knowledge and hybridize with other methods, parallelism, robust to dynamic environments, capability for self –optimization, able to solve problems with no known solutions etc. there are also some other advantages with EC approaches e.g. no need for analytic expression of the problem, no need for derivatives etc.

The Particle Swarm Optimization (PSO) method is a member of the wide category of Swarm Intelligence methods [15], for solving optimization problems. PSO can be easily implemented and it is computationally inexpensive, since its memory and CPU speed requirements are low. Moreover, it does not require gradient information of the objective function under consideration, but only its values, and it uses only primitive mathematical operators. PSO has been proved to be an efficient method for many problems and in some cases it does not suffer the difficulties encountered by other EC techniques.

### 5.1 Evolutionary programming (EP)
EP has been applied with success to many function and combinational optimization problems [13],[14]. Optimization by EP can be summarized into two major steps:
(1) mutate the solutions in the current population,
(2) select the individuals for the next generation from the mutated and current solutions.
These two steps are a population-based version of generate-and-test method, where mutation generates new solutions (offspring) and a selection test that newly generates new solutions should survive to the next generation. Generate –and-test framework can be defined as shown in fig .1. Various models of mutation have been given, the most successful form are (a) gaussian mutation (b) cauchy mutation (more generalize form is Levy distribution) (c) mixed mutation including gaussian and cauchy mutation. In the gaussian mutation the search step is sometimes not large enough for the individual to jump out the local optimum hence cauchy mutation can be taken as solution to solve this problem.

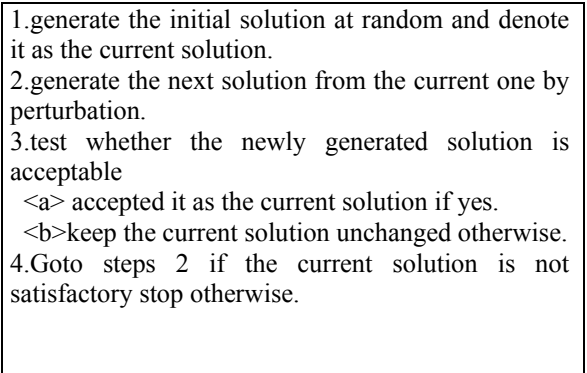

Fig.1.Generate and Test framework

Nevertheless a large search step size may not be beneficial at all if the current search point is already very close to the global optimum. A remedy of above defined problem associated with gaussian and cauchy mutation, a hybrid approach having facility to generate an offspring from a parent by gaussian and cauchy mutation separately and among the two offspring, one which will have high value of fitness will accepted as survived offspring for that parent

## 6 Particle Swarm Optimization
### 6.1 The Particle Swarm Optimization algorithm

PSO's precursor was a simulator of social behavior that was used to visualize the movement of a birds' flock. Several versions of the simulation model were developed [15],[16],[17], incorporating concepts such as nearest-neighbor velocity matching and acceleration by distance . When it was realized that the simulation could be used as an optimizer, several parameters were omitted, through a trial and error process, resulting in the first simple version of PSO. PSO is similar to EC techniques in that, a population of potential solutions to the problem under consideration is used to probe the search space. However, in PSO, each individual of the population has an *adaptable velocity* (position change), according to which it moves in the search space. Moreover, each individual has a *memory*, remembering the best position of the search space it has ever visited. Thus, its movement is an aggregated acceleration towards its best previously visited position and towards the best individual of a topological neighborhood. Two variants of the PSO algorithm were developed. One with a global neighborhood, and one with a local neighborhood. According to the global variant, each particle moves towards its best previous position and towards the best particle in the whole swarm. On the other hand, according to the local variant, each particle moves towards its best previous position and towards the best particle in its restricted neighborhood. In the following paragraphs, the global variant is exposed (the local variant can be easily derived through minor changes). Suppose that the search space is $D$-dimensional, then the $i$-th particle of the swarm can be represented by a $D$-dimensional vector, $X_i = [x_{i1}, x_{i2}, \ldots x_{iD}]$. The *velocity* (position change) of this particle can be represented by another $D$-dimensional vector $V_i = [v_{i1}, v_{i2}, \ldots v_{iD}]$. The best previously visited position of the $i$-th particle is denoted as $P_i = [p_{i1}, p_{i2}, \ldots p_{iD}]$. Defining '$g$' as the index of the best particle in the swarm (i.e., the $g$-th particle is the best),'n' is the best seen by that particular particle and let the superscripts denote the iteration number, then the swarm is manipulated according to the following two equations [16]

$$V(n+1)_{id} = \chi [w V_{nid} + C_1 r1 (P_{nid} - X_{nid}) + C_2 r2 (P_{ngd} - X_{nid})] \quad \ldots (2)$$

$$X(n+1)_{id} = X_{nid} + V(n+1)_{id} \quad \ldots (3)$$

Where $w$ is called *inertia weight*; $c1$, $c2$ are two positive constants, called *cognitive* and *social* parameter respectively; and $\chi$ is a *constriction factor*. The role of these parameters is discussed in the next section. In the local variant of PSO, each particle moves towards the best particle of its neighborhood.

Indeed, the swarm in PSO performs space calculations for several time steps. It responds to the quality factors implied by each particle's best position and the best particle in the swarm, allocating the responses in a way that ensures diversity. Moreover, the swarm alters its behavior (state) only when the best particle in the swarm (or in the neighborhood, in the local variant of PSO) changes, thus, it is both adaptive and stable.

### 6.2 The parameters of PSO

The role of the *inertia weight w*, in Equation (2), is considered critical for the PSO's convergence behavior. The inertia weight is employed to control the impact of the previous history of velocities on the current one. Accordingly, the parameter $w$ regulates the trade-off between the global (wide-ranging) and local (nearby) exploration abilities of the swarm. A large inertia weight facilitates global exploration (searching new areas), while a small one tends to facilitate local exploration, i.e., fine-tuning the current search area. A suitable value for the inertia weight $w$ usually provides balance between global and local exploration abilities and consequently results in a reduction of the number of iterations required to locate the optimum solution.

Initially, the inertia weight was constant. However, experimental results indicated that it is better to initially set the inertia to a large value, in order to promote global exploration of the search space, and gradually decrease it to get more refined solutions. Thus, an initial value around 1.2 and a gradual decline towards 0 can be considered as a good choice for *w*. The parameters *c*1 and *c*2, in Equation (2), are not critical for PSO's convergence. However, proper fine-tuning may result in faster convergence and alleviation of local minima. An extended study of the acceleration parameter in the first version of PSO is given in (Kennedy, 1998). As default values, *c*1 = *c*2 = 2 were proposed, but experimental results indicate that *c*1 =*c*2 = 0.5 might provide even better results. Recent work reports that it might be even better to choose a larger cognitive parameter, *c*1, than a social parameter, *c2*, but with *c*1 + *c*2 ≥4 (Carlisle and Dozier, 2001). The parameters *r*1 and *r*2 are used to maintain the diversity of the population, and they are uniformly distributed in the range [0, 1]. The constriction factor $\chi$ controls on the magnitude of the velocities, in a way similar to the *Vmax* parameter, result in a variant of PSO, different from the one with the inertia weight.

### 6.4 Differences between PSO and EC techniques

In EC techniques, three main operators are involved. The *recombination*, the *mutation* and the *selection* operator. PSO does not have a direct recombination operator. However, the stochastic acceleration of a particle towards its previous best position, as well as towards the best particle of the swarm (or towards the best in its neighborhood in the local version), resembles the recombination procedure in EC .In PSO the information exchange takes place only among the particle's own experience and the experience of the best particle in the swarm, instead of being carried from fitness dependent selected "parents" to descendants as in GA's. Moreover, PSO's directional position updating operation resembles mutation of GA, with a kind of memory built in. This mutation-like procedure is multidirectional both in PSO and GA, and it includes control of the mutation's severity, utilizing factors such as the *Vmax* and $\chi$.PSO is actually the only evolutionary algorithm that does not use the "survival of the fittest" concept. It does not utilize a direct selection function. Thus, particles with lower fitness can survive during the optimization and potentially visit any point of the search space.

### 7 EXPERIMENTAL SETUP FOR COMPARATIVE ANALYSIS BETWEEN EP AND PSO.

For network shown in fig.7, assuming the current total load in the network is 60%of total capacity. With different variations of EP and PSO as shown in table (2), task has defined for each case as: for what distribution of flow the average delay will be minimum.

### 7.1 Parameter initialization for experiment

In all cases, initial flow distributions defined randomly with uniform distribution in range [1 50]; Population size 150 for Gaussian and Cauchy mutation methods and 100 for Hybrid method (because each parent generate two offspring) and 300 for all variations of PSO, so that next generation is created by a population of 300.For all variation of EP: $\sigma_i$ = 0.01, for all i;
In PSO, $\chi$ = 0.75, w gets changing from 1.2 to 0.1 with iteration wise, C1 and C2 both taken as 0.5.Terminating criteria taken as: when for twenty continuous iteration, if the difference in objective function values having change lese than 0.00000001; For each setup total number of generations taken before termination of execution, total time taken to execute the program and best average delay given at the time of termination by program are consider as final outcomes. The process of execution repeated for ten independent trails and results are shown in table (2). From the results it's very clear that in different variations of EP, gaussian mutation and cauchy mutation based approach nearly perform equally, while hybrid approach, combination of both gaussian and cauchy, perform better comparatively. This is because longer jump of cauchy mutation helps the algorithm at the early stage and smaller step of gaussian, when near to optimal value. But PSO with constraint factor $\chi$ clearly outperforms all other approaches. Performances of this method are shown in fig.2 and in fig.3. Hence for training and test data generation PSO with $\chi$ is selected with same parameters initialization as taken above.

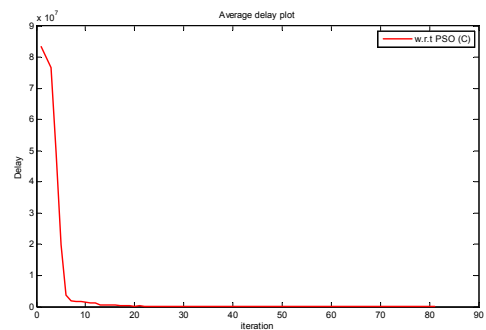

Fig.2.performance on delay minimization

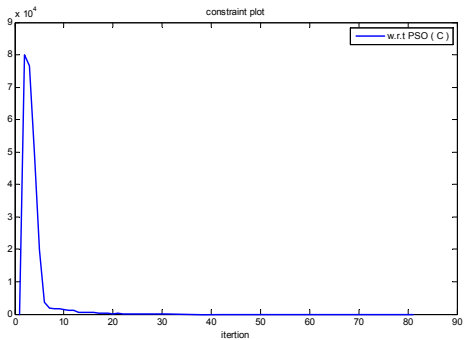

Fig.3.Constrint value performance

## 8 ANN PREDICTIVE MODEL FOR FLOW DISTRIBUTION

Steps taken to develop the model that provide the assistance to routing protocol to define the optimal flow on links have given below and its application with routing protocol shown in fig.4.

(1) Generate the various different possible total load (Li, i=1,2…m) for given network in the range of 30%-95% of the maximum capacity of the network.
(2) For each value of Li find the optimal value of flow vector f, using an optimization method.
(3) Create a data set containing number of possible load and its corresponding optimal flow distribution, this is training data set.
(4) Create a data set as in step 3 with remaining information in step 2, this is test data set.
(5) Give training to neural network with training data set, by taking input as the total load and its corresponding optimal flow distribution as target.
(6) Verify the result with test data set.

Fig.4. Steps of predictive model for flow distribution

The role of ANN is to provide the optimal distribution of flow for any present load in the network. This distribution is a very useful information for routing protocol for assigning the flow on each link so that minimum average delay with given load can achieve. Any variation in total load of traffic matrix will sense by neural network and in result immediate optimal distribution. Hence neural network can be use to provide the information of flow distribution for any variation of load that may happen with network, very quickly and precisely.

Training data and test data set for various total loads, the optimum distribution, which could generate minimum delay, has shown in table (3) and in table (4). Along with that, Mean link utilization (M.L.U) and number of generation taken has also given. Training has given over training dataset with the feed forward architecture and a variation of back propagation learning algorithm [18]. Specification of architecture and learning algorithm has given below:

Size of architecture: 1 input node, 7 hidden nodes, 13 output nodes; Learning rate =0.9,momentum constant=0.2;Weights are initializing with uniform random distribution in range [-0.5 0.5]; Bias node has applied to all hidden nodes. Total number of allowed iteration =5000;

The curve of learning shown in figure (5).

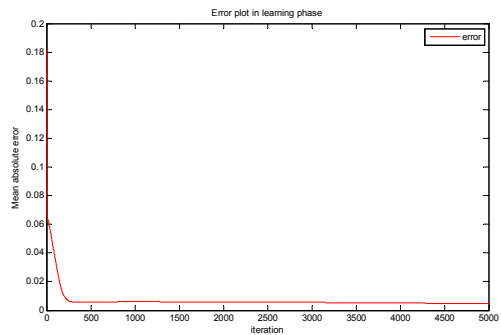

Fig.5.Learning curve of ANN

Test has applied for both training and test data set. Results are shown in table (5) and in table (6). Comparative analysis for training data set and test data set with respect to delay and link utilization has shown in fig.8, 9,10,11.From the numeric values and graphs its very clear the neural network has given the same results as it was expected in both traing and test data set. Hence this trained neural network can applied to predicts the load distribution for any dynamic network, as shown in fig.6.

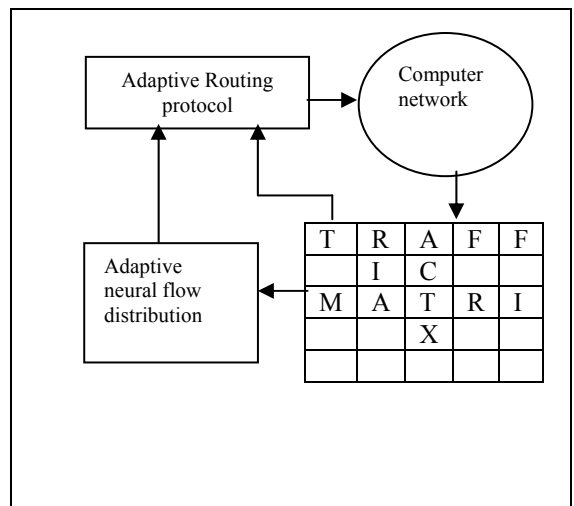

Fig.6.Possible application mode

## EXAMPLE NETWORK

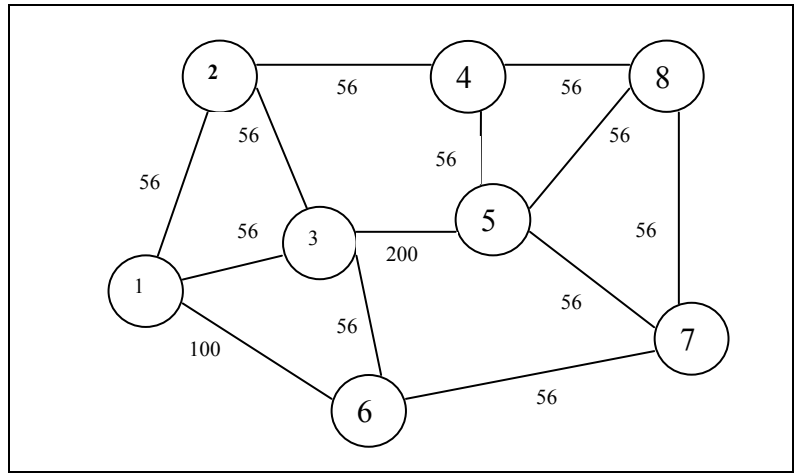

Fig.7. Representing the network taken for simulation, circle represents the nodes and edge connecting the nodes represents the link. Integer number associated with each link represents capacity of that link in (kbps).

Table (1)

| Nodes | 1-2 | 1-3 | 1-6 | 2-4 | 2-3 | 3-5 | 3-6 | 4-5 | 4-8 | 5-8 | 5-7 | 6-7 | 7-8 |
|---|---|---|---|---|---|---|---|---|---|---|---|---|---|
| Link no. | 1 | 2 | 3 | 4 | 5 | 6 | 7 | 8 | 9 | 10 | 11 | 12 | 13 |
| Capacity (kbps) | 56 | 56 | 100 | 56 | 56 | 200 | 56 | 56 | 56 | 56 | 56 | 56 | 56 |

Table (2)
Performances of different variations of EP & PSO to minimize the average delay for a given load. Colum1, Column2 and Column3 are representing total generation, total execution time and minimum delay (msec).

| EP variations ||||||||| PSO variations ||||||
|---|---|---|---|---|---|---|---|---|---|---|---|---|---|---|
| Gaussian ||| Cauchy ||| Hybrid ||| PSO ||| PSO ($\chi$) |||
| 1 | 2 | 3 | 1 | 2 | 3 | 1 | 2 | 3 | 1 | 2 | 3 | 1 | 2 | 3 |
| 345 | 29.6 | 33.3 | 317 | 27.4 | 33.1 | 210 | 13.6 | 33.1 | 427 | 9.6 | 32.7 | 90 | 1.8 | 32.7 |
| 262 | 22.3 | 33.1 | 198 | 17.1 | 33.5 | 291 | 18.7 | 32.7 | 429 | 9.8 | 32.7 | 72 | 1.4 | 32.7 |
| 247 | 21.1 | 35.1 | 211 | 18.2 | 33.2 | 292 | 18.8 | 32.8 | 424 | 9.4 | 32.7 | 62 | 1.2 | 32.7 |
| 261 | 22.2 | 33.6 | 339 | 29.3 | 33.3 | 356 | 22.9 | 32.8 | 425 | 9.4 | 32.7 | 73 | 1.4 | 32.7 |
| 187 | 15.9 | 33.5 | 229 | 19.9 | 33.4 | 347 | 22.4 | 32.8 | 429 | 9.4 | 32.7 | 63 | 1.2 | 32.7 |
| 230 | 19.6 | 33.2 | 282 | 24.4 | 32.9 | 304 | 19.5 | 32.8 | 427 | 9.5 | 32.7 | 92 | 1.8 | 32.7 |
| 292 | 25.0 | 33.6 | 277 | 23.9 | 33.5 | 260 | 16.8 | 32.8 | 425 | 9.4 | 32.7 | 78 | 1.6 | 32.7 |
| 197 | 16.8 | 35.0 | 186 | 16.1 | 32.9 | 215 | 13.8 | 33.0 | 434 | 9.6 | 32.7 | 94 | 1.9 | 32.7 |
| 332 | 28.4 | 33.5 | 271 | 23.4 | 33.2 | 290 | 18.7 | 33.0 | 429 | 9.4 | 32.7 | 60 | 1.2 | 32.7 |
| 220 | 18.8 | 33.7 | 145 | 12.5 | 33.4 | 234 | 15.1 | 32.7 | 435 | 9.7 | 32.7 | 78 | 1.5 | 32.7 |
| Mean ||| Mean ||| Mean ||| Mean ||| Mean |||
| 257 | 22.0 | 33.8 | 245 | 21.2 | 33.2 | 279 | 18.0 | 32.9 | 428 | 9.5 | 32.7 | 76 | 1.5 | 32.7 |

Table(3)
Training data set and its corresponding optimal flow distribution. (M.L.U-Mean link utilization)
(Load represented in Kbps, delay in msec and M.L.U in fractional proportional for all tables)

| Training data set | | | | Flow distribution on links | | | | | | | | | | | | |
|---|---|---|---|---|---|---|---|---|---|---|---|---|---|---|---|---|
| Total Load | Delay | M.L.U | Gen | Link -- 1 | 2 | 3 | 4 | 5 | 6 | 7 | 8 | 9 | 10 | 11 | 12 | 13 |
| 275 | 17.0 | 0.2408 | 67 | 11 | 11 | 40 | 11 | 11 | 114 | 11 | 11 | 11 | 11 | 11 | 11 | 11 |
| 335 | 19.4 | 0.3107 | 65 | 15 | 15 | 46 | 15 | 15 | 123 | 16 | 15 | 15 | 15 | 15 | 15 | 15 |
| 395 | 22.2 | 0.3822 | 67 | 20 | 19 | 51 | 20 | 19 | 131 | 19 | 20 | 19 | 19 | 19 | 19 | 20 |
| 455 | 25.5 | 0.4531 | 73 | 23 | 23 | 57 | 24 | 24 | 139 | 24 | 24 | 24 | 23 | 24 | 23 | 23 |
| 515 | 29.6 | 0.5255 | 77 | 28 | 27 | 62 | 28 | 28 | 146 | 28 | 28 | 28 | 28 | 28 | 28 | 28 |
| 575 | 35.2 | 0.5954 | 89 | 32 | 32 | 68 | 32 | 32 | 155 | 32 | 32 | 32 | 32 | 32 | 32 | 32 |
| 635 | 43.1 | 0.6663 | 88 | 36 | 36 | 74 | 36 | 36 | 163 | 37 | 37 | 36 | 36 | 36 | 36 | 36 |
| 695 | 55.1 | 0.7378 | 84 | 41 | 41 | 79 | 40 | 40 | 171 | 40 | 40 | 41 | 40 | 41 | 40 | 41 |
| 755 | 76.1 | 0.8097 | 75. | 44 | 45 | 85 | 45 | 45 | 178 | 45 | 45 | 44 | 44 | 45 | 45 | 45 |
| 815 | 121.8 | 0.8812 | 110 | 49 | 49 | 90 | 49 | 49 | 186 | 49 | 49 | 49 | 49 | 49 | 49 | 49 |

Table(4)
Test data set and its corresponding optimal flow distribution

| Test Data set | | | | Expected distribution of flow | | | | | | | | | | | | |
|---|---|---|---|---|---|---|---|---|---|---|---|---|---|---|---|---|
| Total Load | Delay | M.L.U | Gen. | Link -- 1 | 2 | 3 | 4 | 5 | 6 | 7 | 8 | 9 | 10 | 11 | 12 | 13 |
| 305 | 18.2 | 0.2753 | 73 | 13 | 13 | 43 | 13 | 13 | 119 | 13 | 13 | 13 | 13 | 13 | 13 | 13 |
| 365 | 20.7 | 0.3462 | 69 | 17 | 17 | 49 | 17 | 17 | 127 | 17 | 17 | 18 | 17 | 17 | 18 | 17 |
| 425 | 23.7 | 0.4176 | 68 | 21 | 22 | 54 | 22 | 22 | 135 | 21 | 21 | 21 | 22 | 21 | 21 | 22 |
| 485 | 27.4 | 0.4901 | 76 | 26 | 26 | 59 | 26 | 25 | 142 | 25 | 26 | 26 | 26 | 26 | 26 | 26 |
| 545 | 32.2 | 0.5610 | 82 | 30 | 30 | 65 | 30 | 30 | 150 | 30 | 30 | 30 | 30 | 30 | 30 | 30 |
| 605 | 38.8 | 0.6309 | 86 | 34 | 34 | 71 | 34 | 34 | 159 | 34 | 35 | 34 | 34 | 34 | 34 | 34 |
| 665 | 48.4 | 0.7018 | 71 | 39 | 38 | 77 | 38 | 38 | 167 | 38 | 39 | 38 | 38 | 39 | 38 | 38 |
| 725 | 64.0 | 0.7742 | 77 | 43 | 43 | 82 | 42 | 43 | 174 | 43 | 42 | 43 | 43 | 42 | 43 | 42 |
| 785 | 94.0 | 0.8451 | 195 | 47 | 47 | 88 | 45 | 47 | 182 | 47 | 47 | 47 | 47 | 47 | 47 | 47 |
| 845 | 173.6 | 0.9156 | 122 | 51 | 51 | 93 | 51 | 51 | 191 | 51 | 51 | 51 | 51 | 51 | 51 | 51 |

Table(5)

| Flow distribution given by ANN on Training data set | | | | | | | | | | | | | | |
|---|---|---|---|---|---|---|---|---|---|---|---|---|---|---|
| Delay | M.L.U | Link-- 1 | 2 | 3 | 4 | 5 | 6 | 7 | 8 | 9 | 10 | 11 | 12 | 13 |
| 19.0 | 0.2673 | 13 | 12 | 41 | 13 | 13 | 113 | 13 | 13 | 12 | 12 | 13 | 13 | 13 |
| 19..9 | 0.3196 | 16 | 15 | 46 | 16 | 16 | 121 | 16 | 16 | 15 | 15 | 16 | 16 | 16 |
| 21.7 | 0.3767 | 19 | 19 | 51 | 19 | 19 | 131 | 19 | 19 | 19 | 19 | 19 | 19 | 19 |
| 24.8 | 0.4445 | 23 | 23 | 56 | 23 | 23 | 140 | 23 | 23 | 23 | 23 | 23 | 23 | 23 |
| 28.7 | 0.5130 | 27 | 27 | 62 | 27 | 27 | 149 | 27 | 27 | 27 | 27 | 27 | 27 | 27 |
| 34.7 | 0.5893 | 31 | 31 | 68 | 31 | 31 | 157 | 32 | 32 | 32 | 32 | 32 | 31 | 32 |
| 42.8 | 0.6640 | 36 | 36 | 74 | 36 | 36 | 164 | 36 | 36 | 36 | 36 | 36 | 36 | 36 |
| 56.5 | 0.7437 | 40 | 41 | 80 | 41 | 41 | 170 | 41 | 41 | 41 | 40 | 41 | 41 | 41 |
| 76.0 | 0.8123 | 45 | 45 | 85 | 45 | 45 | 174 | 45 | 45 | 45 | 45 | 45 | 45 | 45 |
| 115.4 | 0.8777 | 49 | 50 | 90 | 49 | 49 | 177 | 49 | 49 | 49 | 48 | 49 | 49 | 49 |

Table(6 )

| Flow distribution given by ANN on Test data set | | | | | | | | | | | | | | |
|---|---|---|---|---|---|---|---|---|---|---|---|---|---|---|
| Delay | M.L.U | Link-- 1 | 2 | 3 | 4 | 5 | 6 | 7 | 8 | 9 | 10 | 11 | 12 | 13 |
| 19.3 | 0.2924 | 15 | 14 | 43 | 14 | 14 | 117 | 15 | 14 | 14 | 14 | 14 | 14 | 14 |
| 20.6 | 0.3464 | 18 | 17 | 48 | 18 | 17 | 126 | 18 | 17 | 17 | 17 | 17 | 17 | 17 |
| 23.2 | 0.4108 | 21 | 21 | 54 | 21 | 21 | 135 | 21 | 21 | 21 | 21 | 21 | 21 | 21 |
| 26.7 | 0.4789 | 25 | 25 | 59 | 25 | 25 | 145 | 25 | 25 | 25 | 25 | 25 | 25 | 25 |
| 31.4 | 0.5498 | 29 | 29 | 65 | 29 | 29 | 153 | 29 | 30 | 30 | 29 | 29 | 29 | 29 |
| 38.8 | 0.6289 | 33 | 34 | 71 | 34 | 34 | 161 | 34 | 34 | 34 | 34 | 34 | 34 | 34 |
| 48.1 | 0.7004 | 38 | 38 | 77 | 38 | 38 | 167 | 38 | 39 | 38 | 38 | 39 | 38 | 38 |
| 64.9 | 0.7790 | 43 | 43 | 82 | 43 | 43 | 172 | 43 | 43 | 43 | 43 | 43 | 43 | 43 |
| 91.9 | 0.8455 | 47 | 47 | 88 | 47 | 47 | 176 | 47 | 47 | 47 | 47 | 47 | 47 | 47 |
| 160.4 | 0.9112 | 52 | 52 | 92 | 51 | 51 | 178 | 51 | 51 | 51 | 50 | 51 | 51 | 51 |

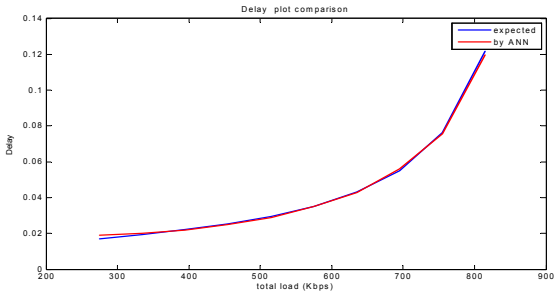

Fig.8.comparative performance on training data for delay

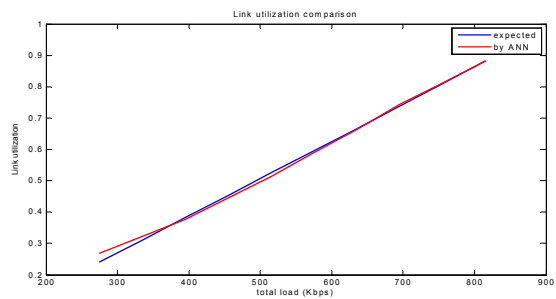

Fig.9.comparative performance on training data for Link utilization

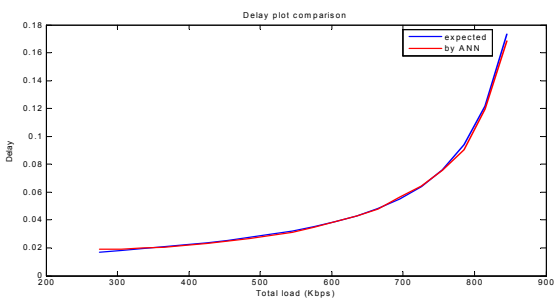

Fig.10.comparative performance on test data for delay

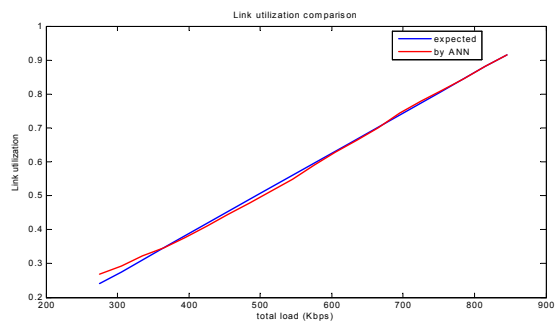

Fig.11.comparative performance on test data for Link utilization

(Blue color represents expected plot and red color given by ANN for fig.8, 9,10,11)

## 9 CONCLUSION

In this paper the challenge to find the optimal flow assignment with dynamic traffic matrix has solved by the application of neural network and particle swarm optimization together. The solution generated by proposed method is very close as it was expected and it's applicable to wide variations of load. To achieve the best performance of system, various variation of evolutionary programming compared with PSO variations. For taken problem, PSO with constriction factor has shown dominancy over other methods. Proposed method is easy to implement and very efficient. It can be integrated with the existing solution with simple engineering to increase the quality of service in terms of minimizing the average delay.

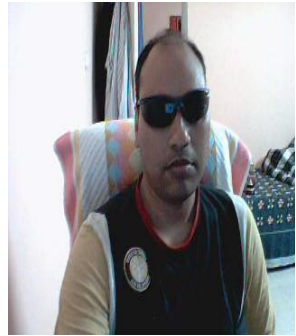

**Manoj kr. Singh is having background of R&D in Nanotechnology,**
**Evolutionary computation, Neural network etc.**
**Currently he is holding the post of director in Manuro Tech. Research. He is also actively associated with industry as a consultant & guiding number of research scholars.**